\documentclass[preprint,12pt]{elsarticle}

\usepackage{hyperref}
\usepackage{algorithm}
\usepackage{algorithmic}
\usepackage{amsmath}
\usepackage{amssymb}

\usepackage{mathtools}
\usepackage{graphicx}
\usepackage{tabularx} 
\usepackage{stfloats}
\usepackage{booktabs} 
\usepackage{makecell}
\usepackage{multirow}
\usepackage{threeparttable}
\usepackage{float}
\usepackage[capitalize,noabbrev]{cleveref}
\usepackage[skip=8pt]{caption}

\journal{Neurocomputing}

\begin{document}

\begin{frontmatter}



\title{Source-Free Domain Adaptive Semantic Segmentation of Remote Sensing Images with Diffusion-Guided Label Enrichment}


\author[label1,label2]{Wenjie Liu}
\author[label1,label2]{Hongmin Liu}
\author[label1,label2]{Lixin Zhang}
\author[label1,label2]{Bin Fan}

\affiliation[label1]{organization={School of Intelligence Science and Technology, University of Science and Technology Beijing}, 
	addressline={30 Xueyuan Road, Haidian District}, 
	city={Beijing},
	postcode={100083}, 
	state={Beijing},
	country={China}}

\affiliation[label2]{organization={Institute of Artificial Intelligence, University of Science and Technology Beijing}, 
	addressline={30 Xueyuan Road, Haidian District}, 
	city={Beijing},
	postcode={100083}, 
	state={Beijing},
	country={China}}

\begin{abstract}

Research on unsupervised domain adaptation (UDA) for semantic segmentation of remote sensing images has been extensively conducted. However, research on how to achieve domain adaptation in practical scenarios where source domain data is inaccessible—namely, source-free domain adaptation (SFDA)—remains limited. Self-training has been widely used in SFDA, which requires obtaining as many high-quality pseudo-labels as possible to train models on target domain data. Most existing methods optimize the entire pseudo-label set to obtain more supervisory information. However, as pseudo-label sets often contain substantial noise, simultaneously optimizing all labels is challenging. This limitation undermines the effectiveness of optimization approaches and thus restricts the performance of self-training. To address this, we propose a novel pseudo-label optimization framework called Diffusion-Guided Label Enrichment (DGLE), which starts from a few easily obtained high-quality pseudo-labels and propagates them to a complete set of pseudo-labels while ensuring the quality of newly generated labels. Firstly, a pseudo-label fusion method based on confidence filtering and super-resolution enhancement is proposed, which utilizes cross-validation of details and contextual information to obtain a small number of high-quality pseudo-labels as initial seeds. Then, we leverage the diffusion model to propagate incomplete seed pseudo-labels with irregular distributions due to its strong denoising capability for randomly distributed noise and powerful modeling capacity for complex distributions, thereby generating complete and high-quality pseudo-labels. This method effectively avoids the difficulty of directly optimizing the complete set of pseudo-labels, significantly improves the quality of pseudo-labels, and thus enhances the model's performance in the target domain. Experimental results demonstrate that DGLE achieves state-of-the-art performance on source-free domain adaptive semantic segmentation benchmarks, including Vaihingen → Potsdam and Rural → Urban for remote sensing, as well as GTA5 → Cityscapes for urban street scenes.

\end{abstract}


\end{frontmatter}

\section{Introduction}
\label{sec1}

Most existing domain adaptation methods for semantic segmentation of remote sensing images \cite{ma2023unsupervised,zhu2023unsupervised,chen2024unsupervised, ma2024decomposition,yan2019triplet,li2022unsupervised} are based on the premise that source domain data is accessible and these methods achieve adaptation through adversarial learning and self-training. However, in many practical scenarios, source domain data is inaccessible due to privacy regulations, storage constraints, or intellectual property restrictions. This complicates knowledge transfer and feature alignment between domains, posing substantial challenges for domain adaptation in semantic segmentation of remote sensing images.

Source-Free Domain Adaptation (SFDA) aims to enable domain adaptation by leveraging a model trained on the source domain and unlabeled target domain data, without requiring access to the source data. Self-training~\cite{zhao2023towards}, as one of the classical strategies for SFDA, generates pseudo-labels to supervise the model's training on the target domain. However, pseudo-labels often contain noise from initial predictions, which can negatively impact model optimization, thus limiting the performance of self-training methods.

\begin{figure}[tb]
    \centering
    \includegraphics[width=1.0\linewidth]{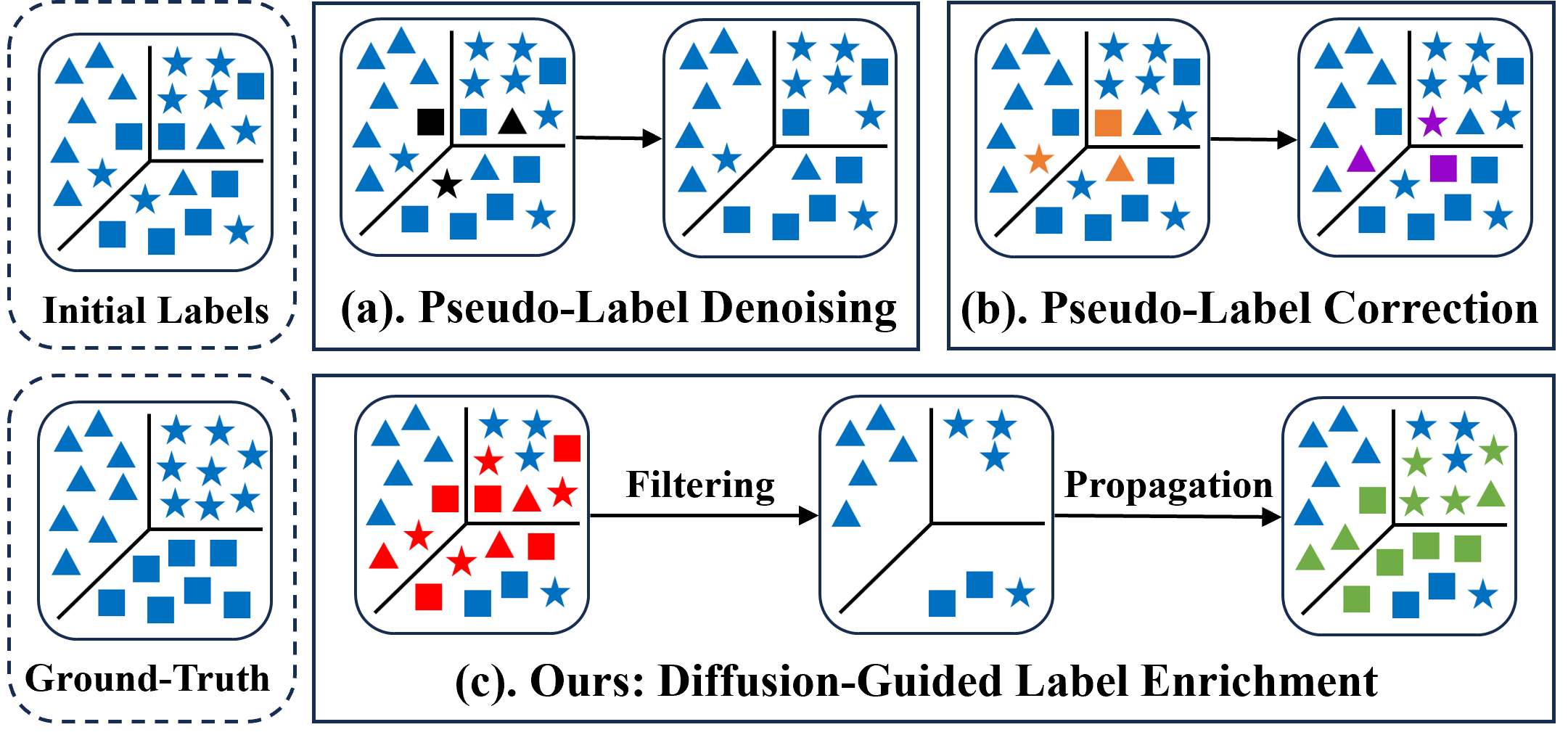}
    \caption{Comparison of different pseudo-label optimization methods: blue represents the initial labels, black represents the labels identified as noise, orange represents the labels identified for correction, purple represents the corrected labels, red represents the labels identified for filtering, and green represents the newly generated labels.}
    \label{fig:0}
\end{figure}

Current research on pseudo-label optimization mainly focuses on denoising and correction, such as confidence-based filtering~\cite{kundu2021generalize} and regional consistency correction~\cite{chen2023semi}. These methods face a trade-off between the quantity and quality of pseudo-labels. To fully utilize the information from the target domain, the optimization typically targets the entire pseudo-label set. However, as the number of pseudo-labels to be optimized increases, the amount of noise within them also increases, thereby increasing the difficulty of optimization and resulting in suboptimal outcomes. We note that obtaining a small number of high-quality pseudo-labels is relatively easy. Based on this insight, we propose a novel pseudo-label optimization framework: Diffusion-Guided Label Enrichment (DGLE). By using a small number of high-quality pseudo-labels as seed and employing diffusion models to propagate them, it is possible to generate a complete set of high-quality pseudo-labels. This framework not only avoids the difficulty of directly optimizing the entire pseudo-label set but also significantly improves the quality of pseudo-labels. Fig.~\ref{fig:0} shows the difference between traditional pseudo-label optimization methods and ours. When compared to existing methods~\cite{toker2024satsynth,lin2025drivegen} that use diffusion models for data generation to increase sample quantity and thereby improve model performance, our proposed DGLE enriches incomplete high-quality pseudo-labels with the contextual features from the input image as a condition, while maintaining the same number of training images.

Our method consists of three stages: \textbf{\textit{1) Seed Pseudo-Label Generation via Pseudo-Label Fusion.}} Although simple confidence filtering can extract reliable information from initial pseudo-labels, the model’s unstable performance in detailed or ambiguous regions may lead to errors. To address this issue, we propose a pseudo-label fusion method based on confidence filtering and super-resolution enhancement to generate seed pseudo-labels. Specifically, we employ an off-the-shelf super-resolution model to enhance the clarity of the target domain images. Considering that super-resolution enhancement may cause some loss of contextual information from the original image and thereby affect the overall accuracy of the pseudo-labels, we introduce a cross-validation mechanism. Pseudo-labels are generated and filtered separately for both original and enhanced images, and their intersection is taken as the seed pseudo-labels. This method leverages super-resolution to mitigate the segmentation model’s shortcomings in detail and boundary processing, while simultaneously incorporating the contextual information of the original image for cross-validation, ensuring the high quality of the seed pseudo-labels. \textbf{\textit{2) Pseudo-Label Propagation with Diffusion Model.}} The information in seed pseudo-labels is incomplete and irregularly distributed, and propagating the seed pseudo-labels while ensuring the quality of newly generated labels is challenging. Diffusion models possess a strong ability to remove noise from random distributions. They can effectively deal with irregular information distribution in seed pseudo-labels. Their powerful ability to model complex distributions and their progressive generation mechanism can accurately capture the semantic distribution and structural features of the seed pseudo-labels, generating accurate predictions. Motivated by these advantages, we propose a diffusion-based pseudo-label propagation framework. In this framework, the contextual features of the input image are used as conditions, and seed pseudo-labels serve as the supervisory signals for training the diffusion model within a ``Noise-to-Map'' generation paradigm. For regions with labels, the model is able to leverage label supervision and contextual features to recover accurate predictions from noise. As training progresses, the model improves its denoising capability, enabling it to achieve satisfactory results even in unlabeled regions. Consequently, the trained model can progressively generate complete and high-quality pseudo-labels during the inference stage. \textbf{\textit{3) Target Domain Model Refinement.}} We use the high-quality pseudo-labels generated by the diffusion model as supervisory signals and apply cross-entropy loss to fine-tune the segmentation model, thereby improving performance in the target domain. To sum up, the main contributions of this paper are as follows:

\begin{enumerate}
\item A new framework, called Diffusion-Guided Label Enrichment (DGLE), is proposed for pseudo-label optimization in source-free domain adaptation. In this framework, pseudo-labels are first filtered to eliminate low-quality samples. Then, a diffusion model is leveraged to propagate the filtered pseudo-labels, resulting in a complete set of high-quality pseudo-labels. This framework effectively avoids the challenges encountered in directly optimizing the entire pseudo-label set.

\item We propose a pseudo-label fusion method that combines confidence filtering and super-resolution enhancement, utilizing cross-validation of details and contextual information to reduce noise in pseudo-labels.

\item Our method achieves new state-of-the-art performance on the SFDA tasks for Vaihingen → Potsdam, Rural → Urban, and GTA5 → Cityscapes, surpassing the previous bests by $\textbf{2.45\%}$, $\textbf{2.18\%}$ and $\textbf{2.5\%}$, respectively.

\end{enumerate}

The remainder of this paper is organized as follows. Section II introduces the related work. Section III provides a detailed introduction to our proposed method. Section IV presents experimental validations to assess the performance of the proposed method. Finally, Section V concludes this work.

\section{Related Work}

\subsection{Unsupervised Domain Adaptation} Unsupervised Domain Adaptation (UDA) aims to improve model performance on the target domain by leveraging labeled data from the source domain and unlabeled data from the target domain. The core challenge in UDA is to address the distributional discrepancy between the source and target domains, commonly referred to as domain shift or domain gap. Adversarial learning and self-training are currently the predominant approaches in this field.

Adversarial learning~\cite{vu2019advent,du2021cross,lu2020stochastic} introduces a domain discriminator to guide the model to learn features that are indistinguishable between the source and target domains. This process aligns feature distributions and mitigates domain discrepancies. For instance, Benjdira et al.~\cite{benjdira2019unsupervised} were among the first to apply adversarial learning in this field. TriADA~\cite{yan2019triplet} proposed triplet adversarial domain adaptation to further improve generalization. To enhance high-level feature extraction, some methods incorporate attention mechanisms. MASNet~\cite{zhu2023unsupervised} employed adversarial learning in the output space and proposed a category-attention-driven prototype memory module. MBATA-GAN~\cite{ma2023unsupervised} utilized self-attention and cross-attention for global feature alignment between domains, and used a multi-discriminator module to learn both domain-invariant and domain-specific features. More recently, De-GLGAN~\cite{ma2024decomposition} combined high-low frequency feature decomposition with a global-local adversarial network to align global and local features across multiple scales.

Self-training~\cite{hoyer2022daformer,hoyer2022hrda,zhao2024unsupervised} utilizes high-confidence predictions from a source model on the target domain to generate pseudo-labels. These pseudo-labels subsequently guide the model’s training on target domain, facilitating adaptation from the source to the target domain. For example, Li et al.~\cite{li2022unsupervised} proposed a Transformer-based framework with Gradual Class Weights and Local Dynamic Quality to improve pseudo-labels quality. LoveCS~\cite{wang2022cross} introduced multi-scale pseudo-label generation, enhancing quality and diversity of pseudo-labels via multi-scale processing and class-balanced sampling. Additionally, STADA~\cite{liang2023unsupervised} combined feature-level adversarial alignment with conditional self-training, and DCA~\cite{wu2022deep} integrated self-training and deep covariance alignment, generating pseudo-labels in stages and explicitly aligning category features.

\subsection{Source-Free Domain Adaptation} Although UDA methods have substantially improved target domain performance without requiring labeled target data, source domain data may be inaccessible during the adaptation process in real-world scenarios, due to the concerns such as data privacy or transfer costs. In response to these challenges, a new paradigm known as Source-Free Domain Adaptation (SFDA) has emerged, which aims to achieve domain adaptation using only a model pretrained on source data and unlabeled target domain data.

Early SFDA methods primarily focused on classification tasks using techniques such as feature alignment~\cite{ding2022source,xia2021adaptive}, sample generation~\cite{li2020model}, and self-training~\cite{zhang2021unsupervised,ahmed2022cleaning,liang2020we}. However, semantic segmentation presents greater challenges, as it requires pixel-level classification and places higher demands on the model’s ability for fine-grained representation and modeling. Most SFDA methods for semantic segmentation rely on self-training~\cite{liu2021source}, with particular emphasis on the effective utilization of pseudo-labels. These approaches focus on denoising and optimizing pseudo-labels to provide higher-quality supervision when adapting the source model to the target domain. SFDASEG~\cite{kundu2021generalize} introduced a multi-head architecture to assist in pseudo-label generation and employed a conditional prior-enforcing auto-encoder to further improve pseudo-label quality. DT-ST~\cite{zhao2023towards} proposed a dynamic teacher update mechanism and a training consistency-based resampling strategy, which together significantly enhance the stability of self-training as well as the quality of pseudo-labels. CROTS~\cite{luo2024crots} alleviated class imbalance and effectively reduced pseudo-label noise by incorporating spatial-aware data mixing and rare-class patch mining. More recently, SND~\cite{zhao2024stable} filtered pseudo-labels based on prediction stability and denoised them via stable neighbor retrieval and category compensation, thereby further improving pseudo-label quality.

There are very few methods addressing SFDA for semantic segmentation of remote sensing images. To date, only the APD~\cite{gao2024attention} has been proposed. APD significantly enhanced segmentation performance in the target domain by leveraging attention-guided prompt tuning and pseudo-label-based feature alignment. In this paper, we adopt the self-training framework and propose a Diffusion-Guided Pseudo-Label Enrichment method to achieve SFDA for semantic segmentation of remote sensing images.

\subsection{Diffusion Model} The diffusion model~\cite{ho2020denoising} has emerged in recent years as a significant breakthrough in deep learning for data generation. Its core idea is to generate data through a progressive denoising process. Numerous methods leverage diffusion models to generate diverse data, thereby improving model performance by increasing the amount of training data. For instance, SatSynth~\cite{toker2024satsynth} modeled the joint distribution of images and labels to synthesize new image-mask pairs, enriching remote sensing semantic segmentation datasets and improving model generalization. Similarly, Dataset Diffusion~\cite{nguyen2023dataset} utilized text-to-image diffusion models and innovative attention mechanisms to automatically generate high-quality semantic segmentation data without manual annotation. FreeMask~\cite{yang2023freemask} employed diffusion models to automatically synthesize segmentation data conditioned on semantic masks, and improved training performance through noise filtering and resampling.

Diffusion models have also found broad applications in SFDA, where existing methods achieve domain adaptation by leveraging the powerful data generation capabilities of diffusion models. For example, DM-SFDA~\cite{chopra2024source} leveraged the confidence scores predicted by the source model to guide the diffusion model in generating source-like data, and then performed mixed training with target domain data, thereby narrowing the gap between the source and target domains. RKP~\cite{zang2024generalized} integrated text-to-image diffusion models with pseudo-labeling and multi-domain style control to synthesize diverse data exhibiting both target and cross-domain styles, thereby improving model performance not only in the target domain but also across source and previously unseen domains. Different from these methods, our proposed DGLE focuses on completing incomplete pseudo-labels, rather than generating new data to augment the training set.

\section{Methodology}
\subsection{Preliminary}

\begin{figure*}[tb]
    \centering  
    \includegraphics[width=1\linewidth]{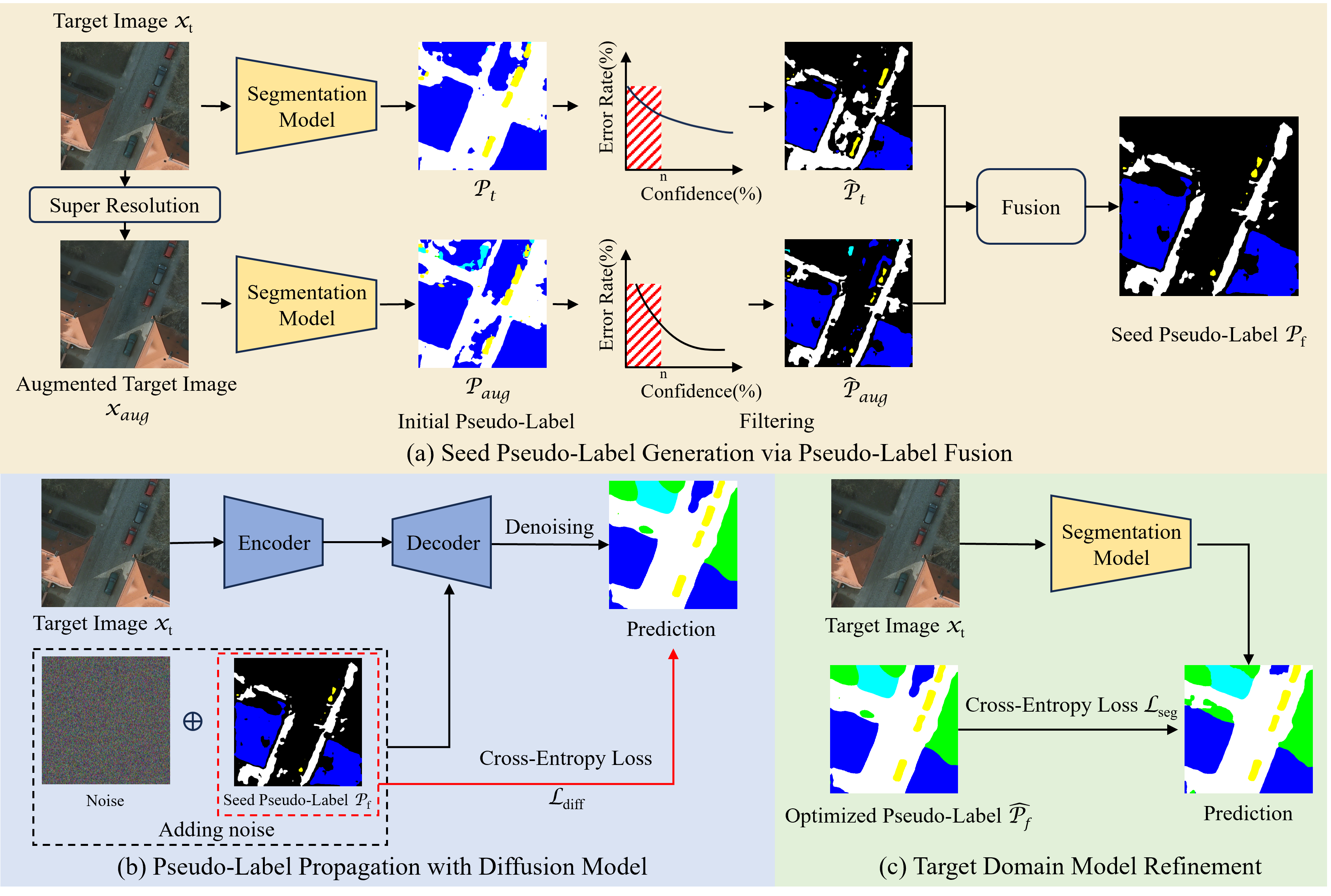}  
    \caption{The pipeline of DGLE: (a) Seed Pseudo-Label Generation via Pseudo-Label Fusion: Target domain images and their super-resolved versions produced by REAL-ESRGAN, are input to the segmentation model to generate initial pseudo-labels. These pseudo-labels are then filtered according to confidence scores and fused to create high-quality seed pseudo-labels. (b) Pseudo-Label Propagation with Diffusion Model: The seed pseudo-labels serve as supervisory signals to train the diffusion model. The trained model is then used for inference to produce high-quality pseudo-labels. (Note: $\mathcal{P}_f$ is only used during training.) (c) Target Domain Model Refinement: The segmentation model is further refined using the high-quality pseudo-labels with cross-entropy loss.}  
    \label{fig:1}

\end{figure*}

In the setting of SFDA, we are provided with a segmentation model $\mathcal{G}(\theta)$ that has been trained on a labeled source domain dataset. The goal is to adapt the model $\mathcal{G}(\theta)$ so that it can achieve good performance on the unlabeled target domain dataset $\mathcal{D}_t = \{x_t^i\}_{i=1}^{N_t}$ without accessing the source dataset. To achieve this, our method adopts self-training strategies to optimize the model $\mathcal{G}(\theta)$. The loss function to be minimized is as follows:
\begin{equation}  
\mathcal L = \sum_{i}^{N_t} \sum_{l}^{H \times W} \mathcal{L} \big[ \mathcal{G}(x_t^{(i,l)} \mid \theta), \hat{y}_t^{(i,l)} \big], 
\label{eq:loss} 
\end{equation}
where $H$ and $W$ are the width and height of the image, $i$ denotes the index of the $i$-th image in the target domain dataset, $N_t$ is the number of images in the target domain dataset, $\mathcal{L}$ is the cross-entropy loss and $\hat{y}_t$ is the pseudo-labels. Our goal is to improve the quality of pseudo-labels $\hat{y}_t$, thereby enhancing the performance of the model $\mathcal{G}(\theta)$. Fig.~\ref{fig:1} shows the pipeline of the proposed DGLE method, which consists of three main steps. First, seed pseudo-labels are generated via pseudo-label fusion. Next, the seed pseudo-labels are propagated using diffusion models to obtain complete and high-quality pseudo-labels. Finally, the target domain model is refined with the propagated pseudo-labels to improve performance.

\subsection{Seed Pseudo-Label Generation via Pseudo-Label Fusion}  

To generate high-quality seed pseudo-labels, we propose a pseudo-label fusion method based on confidence filtering and super-resolution enhancement. Specifically, we utilize the pre-trained super-resolution enhancement model Real-ESRGAN~\cite{wang2021real} to enhance the target domain data $\mathcal{D}_t = \{x_t^i\}_{i=1}^{N_t}$, resulting in an enhanced dataset $\mathcal{D}_{aug}$:  
\begin{equation}   
\mathcal{D}_{aug} = \{x_{aug}^i \mid x_{aug}^i = SR(x_t^i), x_t^i \in \mathcal{D}_t\},  
\end{equation}  
where $\text{SR}(\cdot)$ denotes the super-resolution enhancement function and $\mathcal{D}_{aug}$ represents the enhanced dataset obtained by applying $\text{SR}$ to each sample in $\mathcal{D}_t$. 

Next, we utilize the segmentation model $\mathcal{G}(\theta)$ to perform inference on both the target domain data $\mathcal{D}_t$ and the augmented data $\mathcal{D}_{aug}$, generating two initial pseudo-label sets. For the target domain data, the segmentation model $\mathcal{G}(\theta)$ predicts the semantic labels $y_t^i$ and the confidence scores $\mathcal{C}_t^i$ for each pixel. The confidence score $\mathcal{C}_t^i$ denotes the predicted probability (obtained via the softmax function) for each pixel's assigned class. The pseudo-labels $\mathcal{P}_t$ for the target domain data can be expressed as:  
\begin{equation}   
\mathcal{P}_t = \{(y_t^i, \mathcal{C}_t^i)\}_{i=1}^{N_t},  
\end{equation}  
Similarly, for the augmented data $\mathcal{D}_{aug}$, the model predicts the semantic labels $y_{aug}^i$ and confidence scores $\mathcal{C}_{aug}^i$, which constitutes another pseudo-label set $\mathcal{P}_{aug}$, expressed as follows:  
\begin{equation}   
\mathcal{P}_{aug} = \{(y_{aug}^i, \mathcal{C}_{aug}^i)\}_{i=1}^{N_{aug}}.  
\end{equation}

To obtain reliable pseudo-labels from the initial pseudo-label sets, we perform percentage-based confidence filtering for each category in $\mathcal{P}_t$ and $\mathcal{P}_{aug}$. Since the operations for both datasets are identical, we use $\mathcal{P}_t$ as an example. For each category $c$, we first extract the pixels from the pseudo-labels $\mathcal{P}_t$ that belong to category $c$:  
\begin{equation}  
\mathcal{P}_t^c = \{(y_t^{(i,l)}, \mathcal{C}_t^{(i,l)}) \in \mathcal{P}_t \mid y_t^{(i,l)} = c\},  
\label{eq:5}   
\end{equation}  
where $l \in H \times W$ denotes the pixel position in the image. Next, we extract the corresponding confidence scores of these pixels as follows:  
\begin{equation}  
\mathcal{C}_t^c = \{\mathcal{C}_t^{(i,l)} \mid (y_t^{(i,l)}, \mathcal{C}_t^{(i,l)}) \in \mathcal{P}_t^c\},  
\end{equation}  
The confidence scores are then sorted in ascending order as follows:  
\begin{equation}  
\mathcal{C}_t^{c, sorted} = \text{sort}(\mathcal{C}_t^c, \text{ascending}),  
\end{equation}  
To determine the confidence score threshold $\tau_t^c$ for filtering, we calculate the value corresponding to the specified percentile threshold $n$:  
\begin{equation}  
{\tau}_{t}^{c} = \mathcal{C}_t^{c, sorted}[\lceil n \times |\mathcal{C}_t^{c, sorted}| \rceil],  
\end{equation}  
where $\lceil \cdot \rceil$ denotes the ceiling function, which rounds up to the nearest integer, and \(|\mathcal{C}_t^{c, sorted}|\) is the total number of confidence scores in the set. Finally, we apply this threshold to filter the pseudo-labels, retaining only the pixels with confidence scores at or above the calculated threshold:  
\begin{equation}  
\hat{\mathcal{P}}_t^c = \{ y_t^{(i,l)} \mid y_t^{(i,l)} \in \mathcal P_t^c, \ \mathcal{C}_t^{(i,l)} \geq \tau_t^c \},  
\label{eq:9}   
\end{equation}  
where $\hat{\mathcal{P}}_t^c$ is the filtered pseudo-labels for category $c$. By combining the filtered pseudo-labels of all categories, we obtain the complete filtered pseudo-label set $\hat{\mathcal{P}}_t$.  

Following the same procedure, we filter the pseudo-labels $\mathcal{P}_{aug}$ to obtain $\hat{\mathcal{P}}_{aug}$. We then fuse the two filtered pseudo-label sets $\hat{\mathcal{P}}_t$ and $\hat{\mathcal{P}}_{aug}$ by comparing the predictions on a per-pixel basis. For each pixel, we retain its prediction value only if it is present in both two pseudo-label sets and the predictions are consistent. The seed pseudo-labels $\mathcal{P}_f$ are obtained through this fusion process, expressed as follows:  
\begin{equation}  
\mathcal{P}_f = \left\{ y_t^{(i,l)} \ \middle| \   
\begin{aligned}  
    &y_t^{(i,l)} \in \hat{\mathcal P}_t, \\
    &y_{aug}^{(i,l)} \in \hat{\mathcal P}_{aug}, \\
    &y_t^{(i,l)} = y_{aug}^{(i,l)}  
\end{aligned}  
\right\}.  
\label{eq:fusion}   
\end{equation}

Our approach utilizes percentage-based filtering for each class instead of a simple fixed threshold, which helps to avoid exacerbating class imbalance during filtering. Additionally, by cross-validating the detailed information of the augmented data with the contextual information of the original image, this fusion method effectively reduces potential errors introduced by confidence filtering on individual data instances, thereby improving the quality of the seed pseudo-labels $\mathcal{P}_f$. As a result, $\mathcal{P}_f$ serves as the reliable supervisory signals, playing a crucial role in improving the robustness and stability of subsequent pseudo-label propagation.

\subsection{Pseudo-Label Propagation with Diffusion Model}

For pseudo-label propagation, we construct the diffusion model with reference to the DDP framework~\cite{ji2023ddp} to propagate sparse yet high-quality seed pseudo-labels, ultimately transforming them into complete and high-quality semantic segmentation results.

Specifically, during the training phase, we randomly sample a time step $t$ for each sample and adds noise to the seed pseudo-labels $\mathcal{P}_f$ according to the noise level corresponding to t, resulting in noisy seed pseudo-labels $z_t$. Meanwhile, the input image $x$ is processed by the encoder to extract contextual features, which serve as the condition $c$ to guide the subsequent denoising process. During training, under the supervision provided by the seed pseudo-labels, the diffusion model gradually learns how to leverage the contextual features from the image to recover accurate segmentation predictions for the labeled regions from different noise levels. As training progresses, the model's denoising ability becomes increasingly strong. It can even achieve good prediction results for pure noise under the guidance of the condition. In other words, the model can obtain relatively accurate predictions for unlabeled regions, thereby enabling high-quality propagation of seed pseudo-labels. The output is supervised by the seed pseudo-labels $\mathcal{P}_f$ with cross-entropy loss, ensuring that the model learns to align its predictions with the seed pseudo-labels. The loss is computed only on labeled pixels, expressed as follows:
\begin{equation} 
\mathcal{L}_{diff}(\phi) = \sum_{i}^{N} \mathcal{L} \big[ \mathcal{H}\left(z_t^{i}, c \mid \phi\right), \mathcal{P}_f^{i} \big],
\end{equation}
where $\mathcal{L}$ is the cross-entropy loss, $\mathcal{H}$ denotes the diffusion model’s decoder parameterized by $\phi$, $i$ denotes the index of the $i$-th image in the target domain dataset, $N$ is the number of images in the target domain dataset, $c$ is the condition extracted from the target domain image $x^{i}$ and $z_t$ is the noisy seed pseudo-labels at the randomly sampled time step $t$.

During the inference phase of the diffusion model, the trained model generates the complete semantic segmentation result through a progressive denoising process. The process starts with the initial input $ x_0 $, which is random Gaussian noise. At each iteration step $ t $ (where $ t=1,2,\dots,T $), the model takes the output from the previous step ($ x_{t-1} $), combines it with the condition $ c $ encoded from the original image, and predicts the denoised result for the current step. This iterative process can be expressed as follows:
\begin{equation}
x_t = \mathcal{H}(x_{t-1}, c \mid \phi), \qquad t = 1, 2, ..., T
\label{eq:propagate1} 
\end{equation}
where $ x_0 = \eta \sim \mathcal{N}(0, I) $ is the initial random Gaussian noise, $ c $ denotes the conditional information extracted from the input image by the encoder, $\mathcal{H}(\phi)$ is the decoder of the diffusion model. After $T$ denoising steps, the final prediction for semantic segmentation is obtained as:
\begin{equation}
\hat{\mathcal{P}}_f = x_T.
\label{eq:propagate2} 
\end{equation}

By leveraging the powerful conditional generative and contextual modeling capabilities of diffusion models, our pseudo-label propagation framework effectively transforms sparse and incomplete seed pseudo-labels into complete and high-quality segmentation predictions. The model not only learns to recover accurate segmentation results in labeled regions, but also generalizes to unlabeled regions by exploiting the rich contextual information present in the input image. This diffusion-based propagation mechanism complements the pseudo-labels while ensuring the quality of newly generated labels, thereby laying a solid foundation for subsequent training stages.

\subsection{Target Domain Model Refinement}

After generating complete and high-quality pseudo-labels $\hat{\mathcal{P}}_f$ using the diffusion model, we fine-tune the semantic segmentation model $\mathcal{G}(\theta)$ with these pseudo-labels serving as supervision. The high-quality pseudo-labels $\hat{\mathcal{P}}_f$ provide comprehensive and accurate annotations, thus enabling the model to better adapt to the target domain and improve segmentation performance.

Specifically, the target domain data $\mathcal{D}_t = \{x_t^i\}_{i=1}^{N_t}$ is used as input to the segmentation model $\mathcal{G}(\theta)$, while $\hat{\mathcal{P}}_f$ serves as the supervision signals during training. For each pixel in the target domain, the model predicts its semantic label, which is compared against the pseudo-labels $\hat{\mathcal{P}}_f$ using the cross-entropy loss. The loss function follows the same formulation as Eq.~\ref{eq:loss}, with the pseudo-labels $\hat{y}_t$ replaced by $\hat{\mathcal{P}}_f$.

During training, the model iteratively updates its parameters $\theta$ to minimize the loss, aligning its predictions with the pseudo-labels $\hat{\mathcal{P}}_f$. This process benefits from the completeness and accuracy of $\hat{\mathcal{P}}_f$, which reduces the noise and incorrect supervision that are commonly encountered in traditional self-training methods based on pseudo-labels. As a result, the segmentation model is able to deliver more precise predictions for the target domain and thereby effectively bridges the domain gap.

It is worth noting that the trained segmentation model outperforms the previously described diffusion model used for propagating seed pseudo-labels, in terms of segmentation effectiveness on target domain images. In addition, it has significantly faster inference speed, making it more suitable for practical applications.

Algorithm~\ref{alg:algorithm} summarizes the overall procedure of DGLE. This framework effectively utilizes super-resolution enhancement methods and pseudo-label fusion to improve the quality of seed pseudo-labels and innovatively introduces a diffusion model to propagate incomplete pseudo-labels, significantly enhancing the overall quality of pseudo-labels. The propagated pseudo-labels provide better supervisory information to improve the model's performance in the target domain. 


\begin{algorithm}[H]  
    \caption{Overall procedure of the proposed DGLE}  
    \label{alg:algorithm}
    \textbf{Input:} Unlabeled target dataset $\mathcal{D}_t$; segmentation model $\mathcal{G}(\theta)$ trained in the source domain; pre-trained super-resolution enhancement model SR; decoder of diffusion model $\mathcal{H}(\phi)$.\\
    \textbf{Output:} Segmentation model $\mathcal{G}(\theta)$ for the target domain.  
    
    \begin{algorithmic}[1]  
    
    \STATE \textbf{Stage 1: Seed Pseudo-Label Generation via Pseudo-Label Fusion}  
    \STATE $\mathcal{D}_{aug} = \{\text{SR}(x_t) \mid x_t \in \mathcal{D}_t\}$  
    \STATE Generate initial pseudo-labels: $\mathcal{P}_t = \mathcal{G}(\mathcal{D}_t \mid \theta)$,  $\mathcal{P}_{aug} = \mathcal{G}(\mathcal{D}_{aug}\mid \theta)$.  
    \STATE Filter pseudo-labels by confidence, resulting in $\hat{\mathcal{P}}_t$ and $\hat{\mathcal{P}}_{aug}$ (using Eqs.~\ref{eq:5} -~\ref{eq:9}).  
    \STATE Generate seed pseudo-labels $\mathcal{P}_f$ via pseudo-label fusion (using Eq.~\ref{eq:fusion}).

    \STATE \textbf{Stage 2: Pseudo-Label Propagation with Diffusion Model}  
    \STATE Train diffusion model: 
    \FOR{ $x_t$ in $\mathcal{D}_t$}  
        \STATE Randomly select the time step t
        \STATE Add noise to $\mathcal{P}_f$ according to the noise level associated with $t$ to obtain $z_t$  
        \STATE Extract condition $c$ from $x_t$ using the encoder  
        \STATE $\mathcal{L} = \text{CrossEntropy}\left( \mathcal{H}(z_t, c\mid \phi), \mathcal{P}_f \right)$   

        \STATE Update $\phi$ using the AdamW optimizer 
    \ENDFOR  
    \STATE Generate propagated pseudo-labels $\hat{\mathcal{P}}_f$ (using Eq.~\ref{eq:propagate1} and Eq.~\ref{eq:propagate2}).
    
    \STATE \textbf{Stage 3: Target Domain Model Refinement}  
    \STATE Fine-tune segmentation model $\mathcal{G}(\theta)$:  
    \FOR{$x_t$ in $\mathcal{D}_t$}   
        \STATE $\mathcal{L} = \text{CrossEntropy}(\mathcal{G}(x_t\mid\theta), \hat{\mathcal{P}}_f)$   
        \STATE Update $\theta$ using the SGD optimizer 
    \ENDFOR  
    
    \RETURN Refined segmentation model $\mathcal{G}(\theta)$  
    \end{algorithmic}  
\end{algorithm}

\section{EXPERIMENTS}
\subsection{Datasets and Setup}
\paragraph{Datasets} We conducted experiments on the ISPRS~\cite{rottensteiner2020international} and LoveDA~\cite{wang2021loveda} remote sensing datasets to evaluate the effectiveness of our proposed method. The ISPRS dataset, released by the International Society for Photogrammetry and Remote Sensing (ISPRS), comprises the Potsdam and Vaihingen sub-datasets. The Potsdam images contain four spectral bands—infrared, red, green, and blue (IRRGB) and the Vaihingen images include three spectral bands: infrared, red, and green (IRRG). All images in both Potsdam and Vaihingen are annotated by semantic labels for six ground object categories: Impervious Surface, Building, Low Vegetation, Tree, Car and Clutter. In our experiments, we select Vaihingen IRRG (VH-IRRG) as the source domain dataset and Potsdam RGB (PD-RGB) as the target domain dataset. 

We adopt the same training and evaluation set split as MBATrans~\cite{ma2023unsupervised} for the VH-IRRG and PD-RGB datasets. The VH-IRRG dataset includes 16 VHR true orthophotographs of size 2500 × 2000 with a ground sampling distance (GSD) of 9 cm. We selected 12 images as the training set, corresponding to the orthophotographs indexed by $\{$1, 3, 23, 26, 7, 11, 13, 28, 17, 32, 34, 37$\}$. In contrast, the PD-RGB dataset contains 24 VHR true orthophotographs of size 6000 × 6000 with a GSD of 5 cm. These 24 orthophotographs are split into a training set of 18 images $\{$$6\_10$, $7\_10$, $2\_12$, $3\_11$, $2\_10$, $7\_8$, $5\_10$, $3\_12$, $5\_12$, $7\_11$, $7\_9$, $6\_9$, $7\_7$, $4\_12$, $6\_8$, $6\_12$, $6\_7$, $4\_11$$\}$ and a test set of 6 images $\{$$2\_11$, $3\_10$, $4\_10$, $5\_11$, $6\_11$, $7\_12$$\}$. All original images in VH-IRRG and PD-RGB are divided into patches of 512 × 512 pixels. As a result, a total of 256 images are used for source domain training, 2592 images for target domain training, and 864 images for target domain evaluation.


LoveDA was collected from three cities in China—Nanjing, Changzhou, and Wuhan—and comprises two subsets: Urban and Rural. All images in both subsets are annotated with semantic labels for seven land cover categories: Background, Building, Road, Water, Barren, Forest, and Agriculture. We use Rural as the source domain dataset and Urban as the target domain dataset. All images are of size 1024 × 1024 pixels, with a GSD of 30 cm. Specifically, 1,366 images are used for source domain training, 1,156 for target domain training, and 677 for target domain evaluation.

In addition to remote sensing datasets, we also select the widely used ground-level datasets GTA5~\cite{richter2016playing} and Cityscapes~\cite{cordts2016cityscapes} to further validate the generalizability of our method. Specifically, GTA5 serves as the source domain dataset, while Cityscapes is used as the target domain dataset. The GTA5 dataset comprises 24,966 images with a resolution of 1914 × 1052. The Cityscapes dataset consists of 2,975 training images and 500 validation images, each with a resolution of 2048 × 1024.

\paragraph{Setup} We use DeepLabV2~\cite{chen2017deeplab} with ResNet-101~\cite{he2016deep} as the segmentation network. During the seed pseudo-label generation via pseudo-label fusion, we employ pre-trained Real-ESRGAN~\cite{wang2021real} as the super-resolution enhancement model to enhance the target domain data. For pseudo-label propagation with the diffusion model, we adopt the DDP framework~\cite{ji2023ddp}, where the backbone of the encoder is modified to ResNet-101 to ensure fairness in comparison.

During the training of the segmentation network, we use the SGD optimizer~\cite{bottou2010large} with a momentum of 0.9. The initial learning rate is set to $2.5 \times 10^{-4}$ and is decayed according to a poly policy with a power of 0.9. The batch size is set to 4. The diffusion model is optimized using the AdamW optimizer with an initial learning rate of $6 \times 10^{-5}$ and a weight decay of 0.01. To balance the inference time and performance of the diffusion model, the number of sampling steps during inference is set to 3. Our approach is iterative: after optimizing the base model, the resulting model can be used as a new base model for further optimization, as verified in subsequent experiments. Since the original remote sensing images are divided into small patches, all experiments are conducted on a single RTX 3090 GPU.

\paragraph{Evaluation Metric and Methods for Comparison} We use mean IoU (mIoU) as the evaluation metric. The calculation of IoU is as follows:

\begin{equation}  
\text{IoU} = \frac{\text{TP}}{\text{TP} + \text{FP} + \text{FN}}, 
\end{equation}
where TP, FP, and FN denote the number of true positives, false positives, and false negatives, respectively. 

In the SFDA task for remote sensing data, there is only one publicly published method, namely APD~\cite{gao2024attention}. To enrich the comparative experiments, we also apply several recently released open-source SFDA methods to remote sensing data and compare them with our proposed method, including SFDASEG~\cite{kundu2021generalize}, VPT~\cite{ma2023visual}, and CROTS~\cite{luo2024crots}. For reference, we compare with several recent UDA approaches, including MBATA-GAN~\cite{ma2023unsupervised}, MASNet~\cite{zhu2023unsupervised}, DSSFNet~\cite{chen2024unsupervised}, and De-GLGAN~\cite{ma2024decomposition}. Since UDA methods typically do not calculate the IoU for the clutter class in the PD-RGB dataset, we use $\text{mIoU}_5$ to represent the mIoU of the remaining five classes. 

On the GTA5 → Cityscapes task, we compare our method with recent SFDA methods, including HCL~\cite{huang2021model}, SFDASEG~\cite{kundu2021generalize}, DTST~\cite{zhao2023towards}, CrossMatch~\cite{yin2023crossmatch}, CROTS~\cite{luo2024crots}, SND~\cite{zhao2024stable}, DTST + SND~\cite{zhao2024stable}, VPT~\cite{ma2023visual}, and RKP~\cite{zang2024generalized}.

\subsection{Results and Comparison}
\textit{1) Results on the VH-IRRG → PD-RGB Task:} Table~\ref{tab:results1} presents the results of UDA and SFDA methods alongside our proposed method on the VH-IRRG → PD-RGB task. We also report the ``source only" result, which refers to directly evaluating the source model on the target domain without further adaptation. As shown in Table~\ref{tab:results1}, our method achieves an mIoU of 52.85\%, representing an improvement of 19.67\% over the source-only result and 2.45\% over the second-best SFDA method (CROTS). Compared with UDA methods, the $\text{mIoU}_5$ of our method is only 0.14\% lower than the best UDA approach (De-GLGAN). It is important to note that UDA methods utilize labeled data from the source domain, whereas our source-free setup does not employ any source domain labels. Compared with SFDA methods, our approach achieves much better results. These results demonstrate that pseudo-labels optimized via DGLE can provide high-quality supervisory information for target domain training, thereby enhancing model performance on the target domain.

\begin{figure*}[tb]
    \centering  
    \includegraphics[width=1.0\linewidth]{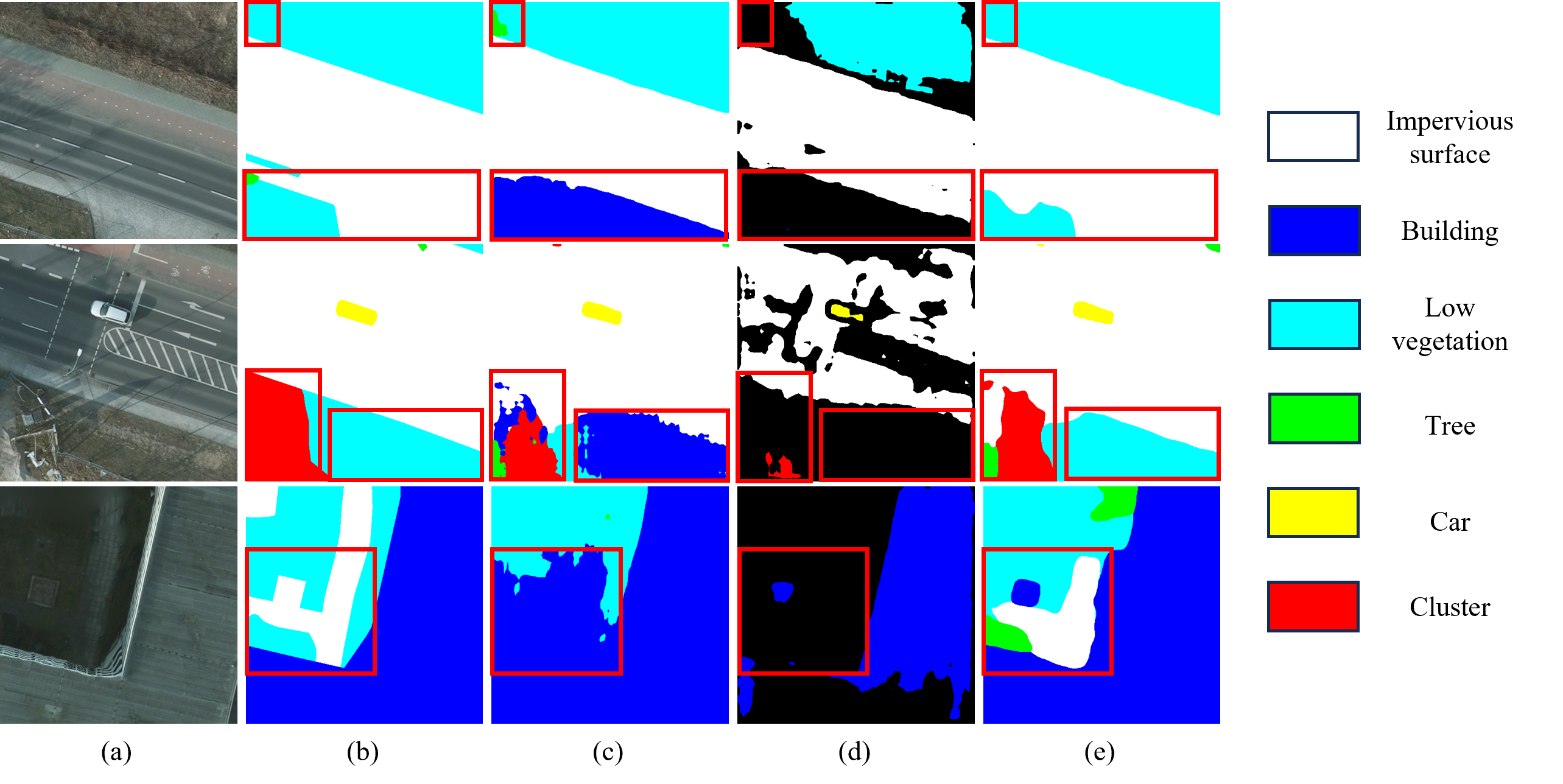}  

    \caption{Qualitative results on the VH-IRRG → PD-RGB task. (a) Target Image. (b) Ground-Truth. (c) Initial Pseudo-Labels. (d) Seed Pseudo-Labels. (e) Propagated Pseudo-Labels.} 
    \label{fig:2} 

\end{figure*}

Additionally, we selected several representative images for visual analysis to further illustrate the advantages of our approach, as shown in Fig.~\ref{fig:2}. It is evident that, although the labeled areas in the seed pseudo-labels are fewer, their accuracy is substantially higher. After propagation using the diffusion model, regions that were previously filtered out are regenerated with high-quality pseudo-labels. The propagated pseudo-labels show significant overall improvement in quality compared to the initial pseudo-labels, which provides further visual evidence of our approach’s effectiveness.

\textit{2) Results on the Rural → Urban Task:} In contrast to the ISPRS dataset, the LoveDA dataset features larger ground sampling distance and exhibits a more pronounced class imbalance between the source and target domains. Consequently, performing the SFDA task on the LoveDA dataset is more challenging and presents greater difficulty. Table~\ref{tab:results2} displays the results on the Rural → Urban task. Our method achieves an mIoU of 46.18\%, which is 12.32\% higher than the source only result and 2.18\% better than the second-best SFDA method (VPT). Although there is a substantial gap between our method and UDA methods due to the greater difficulty of this task compared to the VH-IRRG → PD-RGB task, the results in comparison with SFDA methods further validate the effectiveness of our method.

\begin{figure*}[tb]
    \centering  
    \includegraphics[width=1.0\linewidth]{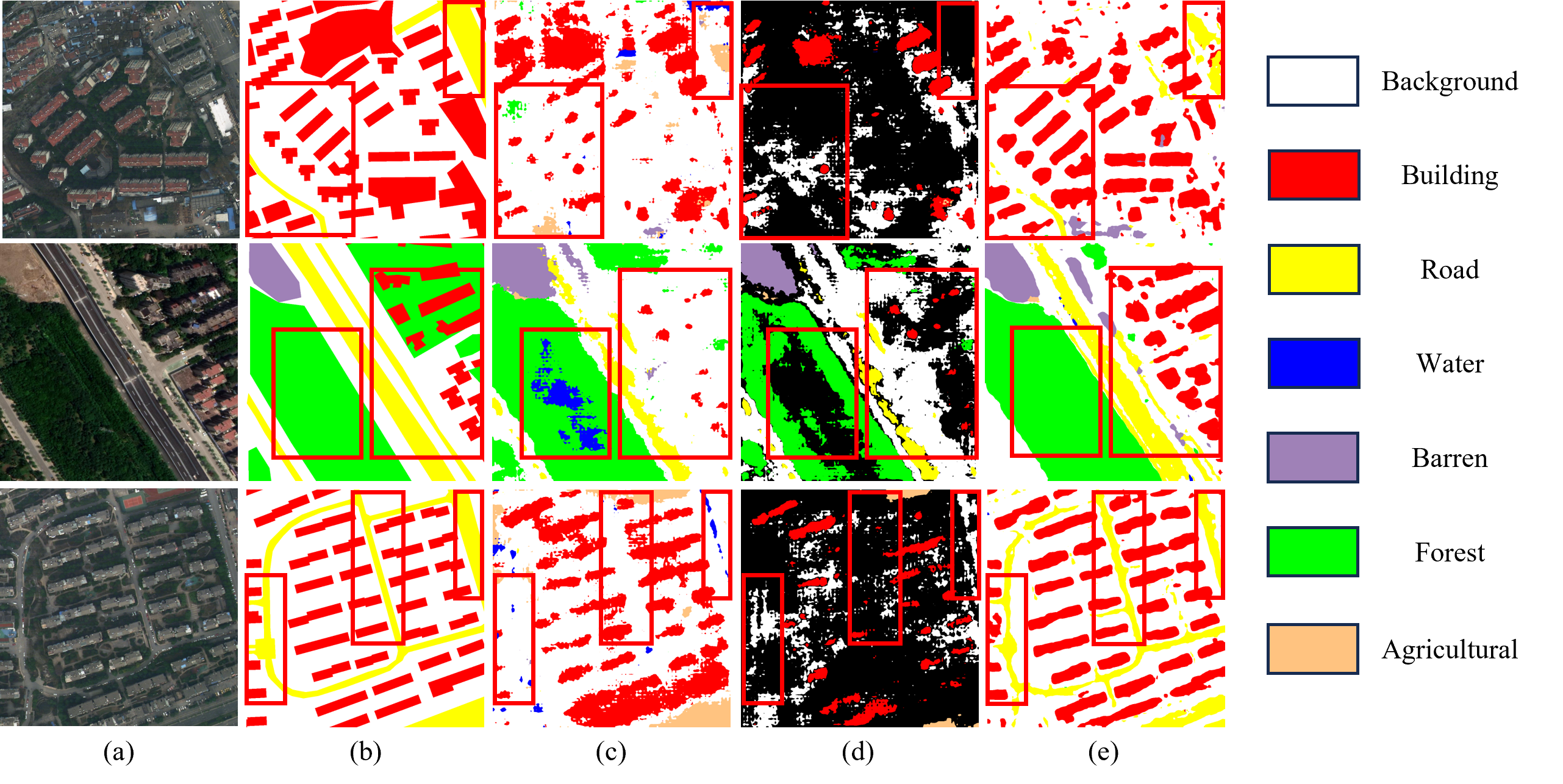}  

    \caption{Qualitative results on the Rural → Urban task. (a) Target Image. (b) Ground-Truth. (c) Initial Pseudo-Labels. (d) Seed Pseudo-Labels. (e) Propagated Pseudo-Labels.} 
    \label{fig:3} 

\end{figure*}

Similarly, we visually demonstrate the effectiveness of our method in optimizing pseudo-labels in Fig.~\ref{fig:3}. By comparing the visualization results across (b),~(c),~and~(e), the improvements achieved by our pseudo-label optimization method can be clearly observed.

\begin{table}[tb]

\caption{Segmentation Results On The VH-IRRG → PD-RGB Task}

\centering  
\resizebox{\textwidth}{!}{
\begin{tabular}{c c c c c c c c c c c}   
\toprule  
{Methods} & Type& {Backbone} & {Imp.} & {Bui.} & {Low.} & {Tre.} & {Car} & {Clu.} & {mIoU} & {$\text{mIoU}_5$} \\ 
\midrule
Source-Only  &- & ResNet101 & 45.49 & 53.15 & 21.6 & 3.21 & 74.33 & 1.33 & 33.18 & 39.55\\
\midrule
MBATA-GAN~\cite{ma2023unsupervised} & \multirow{4}{*}{UDA} & ResNet101 & 75.37 & 81.54 & 58.14 & 30.75 & 69.83 & - & - & 48.32\\
MASNet ~\cite{zhu2023unsupervised}   &  & ResNet101 & 73.97 & 77.89 & 61.89 & 44.87 & 76.04 & - & - & 51.51\\
DSSFNet ~\cite{chen2024unsupervised}  & & ResNet101 & 75.47 & 79.82 & 64.68 & 50.62 & 76.86 & - & - & 54.22\\
De-GLGAN~\cite{ma2024decomposition}  & & Swin-Base & 79.66 & 85.65 & 71.02 & 65.61 & 79.43 & - & - & 62.17\\
\midrule
APD~\cite{gao2024attention}  & \multirow{5}{*}{SFDA}& ViT-H & 60.43 & 61.45 & 43.05 & 40.97 & 62.57 & \textbf{11.26}  & 46.62 & 53.69\\
VPT~\cite{ma2023visual} & & MiT-B5 & \textbf{62.55} & 78.23 & \textbf{47.9} & 34.86 & 59.97 & 0.55 & 47.34 & 56.7\\
SFDASEG~\cite{kundu2021generalize}& & ResNet101 & 61.59 & 78.3 & 29.99 & 41.15 & \textbf{82.2} & 4.5 & 49.62 & 58.64\\

CROTS~\cite{luo2024crots}& & ResNet101 & 55.71 & 76.84 & 40.84 & 46.29 & 76.57 & 6.14 & 50.4 & 59.25\\
 
DGLE (Ours)  & & ResNet101 & 56.92 & \textbf{79.06} & 42.33 & \textbf{53.76} & 78.06 & 6.96  & \textbf{52.85} & \textbf{62.03}\\ 
\bottomrule  
\end{tabular}  
}
\label{tab:results1}
\end{table}

\textit{3) Results on the GTA5 → Cityscapes Task:} To further verify the generality of our proposed DGLE and enable comparison with more SFDA methods, we evaluated our method on the GTA5 → Cityscapes task. As shown in Table~\ref{tab:results}, our method achieves an mIoU of 61.8\% and yields a 2.5\% improvement over the previous best result achieved by RKP, which utilizes diffusion models to generate diverse training data.

\begin{figure*}[tb]
    \centering  
    \includegraphics[width=1.0\linewidth]{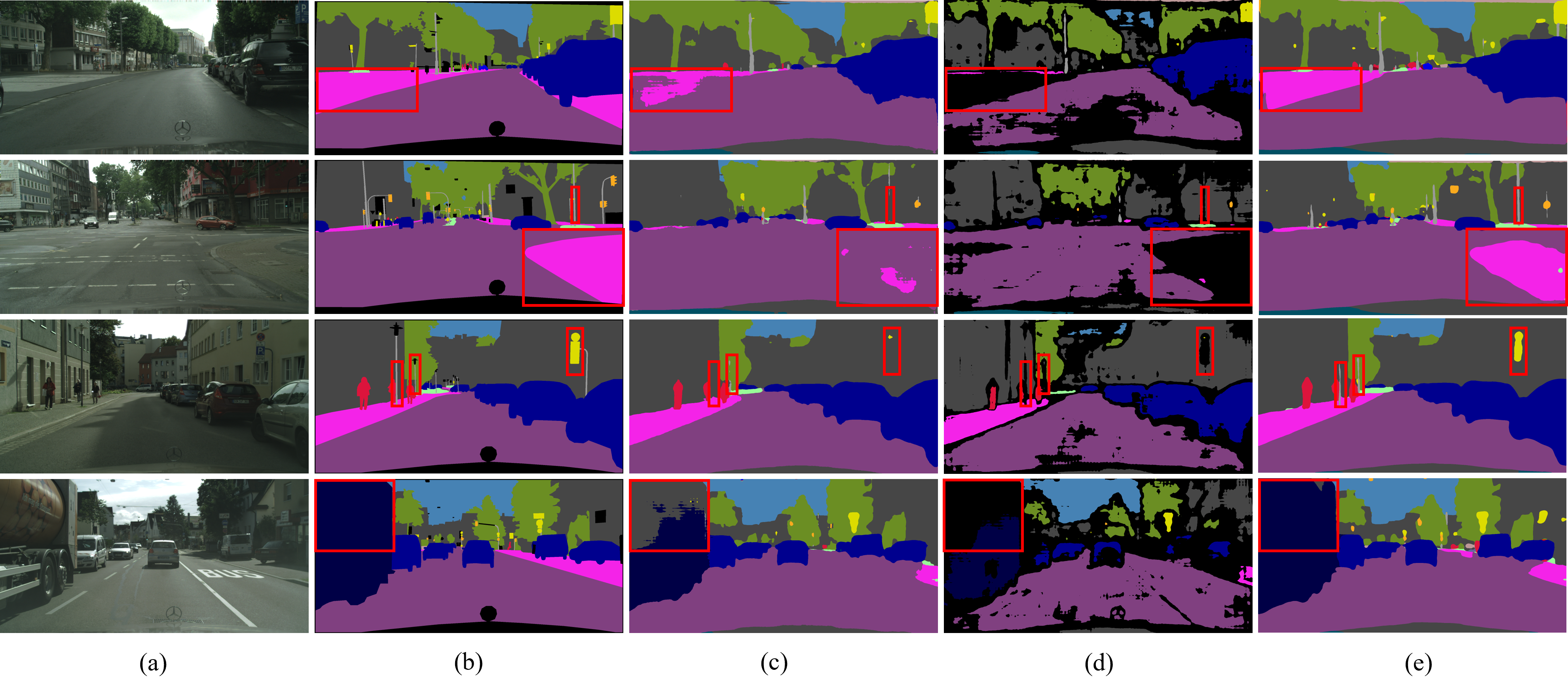}  

    \caption{Qualitative results on the GTA5 → Cityscapes task. (a) Target Image. (b) Ground-Truth. (c) Initial Pseudo-Labels. (d) Seed Pseudo-Labels. (e) Propagated Pseudo-Labels.} 
    \label{fig:4} 

\end{figure*}

Furthermore, we conducted visualization analysis to illustrate the effectiveness of our approach, as shown in Fig.~\ref{fig:4}. By comparing the red-boxed regions of the ground truth, initial pseudo-labels, and propagated pseudo-labels in the third row, it is evident that our method is also highly effective in optimizing pseudo-labels for rare or hard-to-segment classes. This effectiveness can be attributed to our class-wise confidence percentage filtering strategy employed during the seed pseudo-label generation process, which ensures that the retained samples are not dominated by common or easy-to-segment classes, thereby alleviating the class imbalance problem. 

Overall, these results highlight the robustness and strong generalization capability of our method across different datasets and tasks. These findings further demonstrate the practical advantages and general applicability of our approach.

\begin{table}[tb]
\caption{Segmentation Results On The Rural → Urban Task}

\centering  
\resizebox{\textwidth}{!}{
\begin{tabular}{c c c c c c c c c c c}   
\toprule  
Methods &Type& Backbone & Bac. & Bui. & Road & Wat. & Bar. & For. & Agr. & mIoU \\ 
\midrule
Source-Only  & - & ResNet101 & 32.89 & 26.8 & 28.24 & 55.2 & 20.2 & 28.73 & 44.92 & 33.86\\
\midrule
MBATA-GAN~\cite{ma2023unsupervised} & \multirow{4}{*}{UDA} & ResNet101 & 58.66 & 64.95 & 65.90 & 82.54 & 64.35 & 67.72 & 74.18 & 52.52\\
MASNet  ~\cite{zhu2023unsupervised}  &  & ResNet101 & 59.49 & 71.28 & 64.15 & 83.36 & 63.34 & 69.97 & 75.99 & 53.98\\
DSSFNet ~\cite{chen2024unsupervised}  & & ResNet101 & 59.25 & 73.95 & 65.84 & 84.44 & 64.12 & 70.74 & 72.89 & 54.60\\
De-GLGAN~\cite{ma2024decomposition}  & & Swin-Base & 59.53 & 74.49 & 68.80 & 84.66 & 63.71 & 70.50 & 74.83 & 55.50\\
\midrule

SFDASEG~\cite{kundu2021generalize}& \multirow{5}{*}{SFDA}  & ResNet101 & 30.84 & 43.13 & 34.65 & 57.03 & 39.06 & 38.85 & 49.45 & 41.86\\
CROTS~\cite{luo2024crots}& & ResNet101 & 29.43 & \textbf{47.97} & \textbf{47.73} & 64.23 & 35.61 & 43.06 & 28.04 & 42.3\\
APD~\cite{gao2024attention}  &   & ViT-H & \textbf{56.75}  & 44.69  & 38.67  & \textbf{70.31}  & 32.47  & 37.93  & 19.02  & 42.83\\
VPT~\cite{ma2023visual} & & MiT-B5 & 34.45 & 45.1 & 37.78 & 61.51 & \textbf{41.98} & 37.64 & \textbf{49.54} & 44.0\\

DGLE (Ours)  & & ResNet101 & 36.25 & 45.43 & 46.16 & 64.07 & 39.15 & \textbf{43.63} & 48.56 & \textbf{46.18}\\ 
\bottomrule  
\end{tabular}
}
\label{tab:results2}

\end{table}

\begin{table*}[tb]  
\normalsize
\caption{Segmentation Results On The GTA5 → Cityscapes Task}
\centering  
\resizebox{\textwidth}{!}{   
\begin{tabular}{c|c c c c c c c c c c c c c c c c c c c|c }   
\toprule  
 {Method} & \rotatebox{90}{road} & \rotatebox{90}{sidewalk  } & \rotatebox{90}{Building} & \rotatebox{90}{Wall} & \rotatebox{90}{fence} & \rotatebox{90}{pole} & \rotatebox{90}{light} & \rotatebox{90}{sign} & \rotatebox{90}{vege.} & \rotatebox{90}{terrain} & \rotatebox{90}{sky} & \rotatebox{90}{person} & \rotatebox{90}{rider} & \rotatebox{90}{car} & \rotatebox{90}{truck} & \rotatebox{90}{bus} & \rotatebox{90}{train} & \rotatebox{90}{mbike} & \rotatebox{90}{bike} & {mIoU} \\   
\midrule  
HCL~\cite{huang2021model} & 92.6 & 54.6 & 82.8 & 33.2 & 26.2 & 39.8 & 38.1 & 31.9 & 84.5 & 38.6 & 85.3 & 61.3 & 30.2 & 85.4 & 33.1 & 41.6 & 14.4 & 27.3 & 44.0 & 49.7 \\   
SFDASEG~\cite{kundu2021generalize}& 91.7 & 53.4 & 86.1 & 37.6 & 32.1 & 37.4 & 38.2 & 35.6 & 86.7 & \textbf{48.5} & 89.9 & 62.6 & 34.3 & 87.2 & 51.0 & 50.8 & 4.2 & 42.7 & 53.9 & 53.4 \\ 
CROTS~\cite{luo2024crots} & 92.0 & 52.4 & 85.9 & 37.3 & 35.8 & 34.6 & 42.2 & 38.4 & 86.9 & 45.6 & 91.1 & 65.1 & 36.1 & 87.3 & 41.6 & 51.1 & 0.0 & 41.4 & 56.2 & 53.7 \\
VPT~\cite{ma2023visual} & 89.4 & 22.9 & 87.3 & 38.9 & 34.3 & 41.0 & 45.8 & 30.3 & 88.8 & 44.2 & 90.1 & 67.0 & 32.4 & 90.1 & 52.9 & 60.4 & 37.1 & 37.7 & 38.6 & 54.2 \\

DTST~\cite{zhao2023towards} & 93.5 & 57.6 & 84.7 & 36.5 & 25.2 & 33.4 & 44.7 & 36.7 & 86.8 & 42.8 & 81.3 & 62.3 & 37.2 & 88.1 & 48.7 & 50.6 & 35.5 & 48.3 & 59.1 & 55.4 \\
SND~\cite{zhao2024stable} & 93.0 & 54.0 & 84.6 & 35.6 & 30.3 & 31.0 & 41.9 & 41.6 & 87.6 & 44.6 & 86.4 & 62.6 & 38.5 & 87.5 & 48.7 & 42.9 & 36.6 & 49.5 & 58.7 & 55.6 \\
CrossMatch~\cite{yin2023crossmatch} & 95.1 & 67.8 & \textbf{87.7} & \textbf{51.3} & \textbf{41.5} & 36.3 & 47.4 & 51.3 & 87.8 & 47.8 & 87.3 & \textbf{67.0} & 34.2 & 87.5 & 41.0 & 51.8 & 0.0 & 42.6 & 46.4 & 56.4 \\

DTST + SND~\cite{zhao2024stable} & 93.9 & 60.0 & 86.7 & 38.6 & 35.9 & 37.5 & 43.4 & 48.3 & 87.6 & 44.6 & 90.1 & 65.3 & 39.9 & 88.5 & 54.9 & 44.4 & 33.1 & 49.9 & 60.9 & 58.1 \\
RKP~\cite{zang2024generalized} & \textbf{95.9} & 62.2 & 86.9 & 39.6 & 36.9 & 38.7 & 44.4 & 49.0 & \textbf{89.6} & 46.7 & 90.8 & 66.0 & \textbf{41.4} & \textbf{90.3} & 56.0 & 45.3 & 34.5 & 50.5 & 62.1 & 59.3 \\
  
\textbf{DGLE (Ours)} & 95.2 & \textbf{68.9} & 87.5 & 41.2 & 36.1 & \textbf{40.2} & \textbf{48.9} & \textbf{56.5} & 87.3 & 42.5 & \textbf{91.4} & 65.1 & 39.3 & 89.8 & \textbf{66.6} & \textbf{63.2} & \textbf{38.7} & \textbf{53.5} & \textbf{62.5} & \textbf{61.8} \\
\bottomrule  
\end{tabular}  
}  

\label{tab:results}  
\end{table*}

\subsection{Ablation Study}
To validate the effectiveness of each component, we conducted ablation studies on the VH-IRRG $\rightarrow$ PD-RGB task, with the results shown in Table~\ref{tab:2}.

\paragraph{Effectiveness of Pseudo-Label Fusion}
Self-training with confidence-filtered pseudo-labels generated from the original images (Experiment~\#1) achieves an mIoU of $42.53\%$. Similarly, self-training using confidence-filtered pseudo-labels generated from the augmented images (Experiment~\#1') yields an mIoU of $43.03\%$. When fusing the confidence-filtered pseudo-labels from both the original and augmented images for self-training (Experiment~\#2), the mIoU increases to $44.26\%$. These results indicate that the fused pseudo-labels are of higher quality and provide more effective supervisory information, thereby better guiding model training. This demonstrates the feasibility and effectiveness of pseudo-label fusion, which leverages both contextual and detailed information for cross-validation.

\begin{table}[tb] 

    \caption{Ablation Studies on the VH-IRRG → PD-RGB Task.}
    
    \centering
    \resizebox{\textwidth}{!}{ 
    \begin{tabular}{c|c c c c|c}
    \toprule 
    {EXP-ID} & {Confidence Filtering}&{Pseudo-Labels Fusion} & {Pseudo-Labels Propagation} & {Self-Training} & mIoU (\%) \\ \midrule
    \#1 & {\checkmark} & & & {\checkmark} & 42.53\\
    \#1' & {\checkmark} & & & {\checkmark} & 43.03\\
    \#2 & {\checkmark} & {\checkmark} & & {\checkmark} & 44.26\\ 
    \#3 & {\checkmark} & {\checkmark} & {\checkmark} &  & 51.93\\
    \#4 & {\checkmark} & {\checkmark} & {\checkmark} & {\checkmark} & 52.85\\ \bottomrule  
    \end{tabular}
    }
\label{tab:2}
\begin{tablenotes}  
\footnotesize  
\item Note: \#1 only uses original images; \#1' only uses augmented images.  
\end{tablenotes}  
\end{table}

\paragraph{Effectiveness of Pseudo-Label Propagation}
The impact of pseudo-label propagation is evaluated by comparing Experiment~\#2 with Experiment~\#4. Incorporating the pseudo-label propagation method improves the mIoU from $44.26\%$ to $52.85\%$. These ablation results clearly demonstrate the effectiveness of diffusion model-based pseudo-label propagation. Furthermore, as shown in Fig.~\ref{fig:2}(d) and Fig.~\ref{fig:2}(e), we visualized a comparison between the pseudo-labels from Experiment~\#2 (Seed Pseudo-Labels) and Experiment~\#4 (Propagated Pseudo-Labels), both of which were used to train the target domain model. It is evident that after pseudo-label propagation, regions (black areas) previously filtered out in the seed pseudo-labels are regenerated with more accurate segmentation predictions. This further demonstrates that pseudo-label propagation can generate high-quality pseudo-labels, thereby enhancing the model's performance on the target domain.

\paragraph{Effectiveness of Self-Training with Propagated Pseudo-Labels}  

As previously described, self-training with propagated pseudo-labels (Experiment~\#4) achieves higher segmentation accuracy (mIoU $52.85\%$) compared to direct diffusion model inference (Experiment~\#3, mIoU $51.93\%$). In addition, the trained segmentation model offers significantly faster inference speed (29 ms vs. 85 ms per image), making it more suitable for practical applications.


\subsection{Hyper-parameter Analysis}
To assess the impact of various hyper-parameters on our approach, we conducted experiments on the VH-IRRG → PD-RGB task to analyze the influence of several key hyper-parameters, including the number of iterations, the confidence threshold for filtering, and the number of sampling steps used during diffusion model inference.

\paragraph{Number of Iterations} As described in the experimental setup section, our method is iterative: after optimizing the base model, we use the improved model as the new base model for further optimization in subsequent iterations. We evaluated the effect of varying the number of iterations on the experimental results, as shown in Table~\ref{tab:iter}. The model achieves substantial performance gains during the initial iterations. However, as the number of iterations increases, the improvement becomes marginal. Considering the trade-off between performance and computational cost, we set the number of iterations to 4.

\begin{table}[H]  

    \caption{Quantitative Analysis of the number of iterations on the VH-IRRG → PD-RGB Task.}
    \centering  
    \begin{tabular}{c|ccccccc}  
        \toprule  
        iter & 1 & 2 & 3 & 4& 5& 6\\
        \midrule  
        mIoU & 48.07 & 50.55 & 51.46 & 52.85 & 52.86 & 52.88\\
        \bottomrule  
    \end{tabular}
    \label{tab:iter}
\end{table}

\paragraph{Confidence Percentage Thresholds for Filtering}
We analyzed the impact of different confidence percentage thresholds for pseudo-label filtering on the final results. Although increasing the confidence threshold can reduce errors in the pseudo-labels, an excessively high threshold leads to too few pseudo-labels, resulting in limited information and compromising the effectiveness of subsequent pseudo-label propagation. As shown in Table~\ref{tab:n}, setting the confidence percentage threshold to 60$\%$ achieves an effective balance, enabling the model to reach the optimal mIoU of 52.85$\%$.

\begin{table}[H]  

    \caption{Quantitative Analysis of Threshold Filtering Percentage \(n\) on the VH-IRRG → PD-RGB Task.}

    \centering  
    \begin{tabular}{c|cccc}  
        \toprule  
        $n (\%)$ & 20 & 40 & 60 & 80\\
        \midrule  
        mIoU  & 50.72 & 51.92 & \textbf{52.85} & 51.46\\
        \bottomrule  
    \end{tabular}
    \label{tab:n}

\end{table}

\paragraph{Sampling Steps for Diffusion Model Inference} We quantitatively analyzed the effect of the number of sampling steps in the diffusion model, as shown in Table~\ref{tab:sample}. Although increasing the number of sampling steps can improve performance, it also results in longer inference time. Therefore, a balance between inference efficiency and performance is necessary. Based on our experiments, we set the number of sampling steps to 3.

\begin{table}[H]  

    \caption{Quantitative Analysis of the Sampling Steps $T$ of Diffusion Models on the VH-IRRG → PD-RGB Task.}
    \centering  
    \begin{tabular}{c|cccccc}  
        \toprule  
        $T$ & 1 & 2 & 3 & 4& 5 & 6 \\
        \midrule  
        mIoU & 51.41 & 51.92 & 52.85 & 52.87 & 52.79& 52.86\\
        \midrule
        Inference Time (ms) & 58 & 72 & 85 & 103 & 121 & 139 \\
        \bottomrule  
    \end{tabular}
    \begin{tablenotes}  
        \footnotesize  
        \item Note: The reported inference time is the average time required to process a single image. 
    \end{tablenotes} 
    \label{tab:sample}
\end{table}

\section{Conclusion}
In this paper, we propose a pseudo-label optimization framework called Diffusion-Guided Label Enrichment (DGLE) to extract and expand reliable information from seed pseudo-labels.  DGLE first generates a small set of high-quality seed pseudo-labels using a pseudo-label fusion method, and then employs diffusion models to propagate these seed pseudo-labels while ensuring the quality of newly generated labels. This framework circumvents the challenges faced by traditional approaches that attempt to directly optimize the entire pseudo-label set, and achieves state-of-the-art results on source-free domain adaptive semantic segmentation.

\end{document}